\documentclass[sigconf,nonacm]{acmart} %authordraft,anonymous
\usepackage{todonotes}
\usepackage{titlesec}
\usepackage{balance}
\usepackage{graphicx,graphics}

\usepackage{amsmath,amssymb,units}
\usepackage{amsfonts}
\usepackage{cases}
\usepackage{mathrsfs}
\usepackage{multirow}
\usepackage{caption}
\usepackage{color}
\usepackage{flushend}
\usepackage{float}
\usepackage[ruled,vlined,linesnumbered]{algorithm2e}
\usepackage{algorithmic}
\usepackage{subfigure}
\usepackage{extarrows}
\usepackage{array}%tabularx
\usepackage{tabularx}
\usepackage{stmaryrd}
\usepackage{diagbox}
\usepackage{bm}
\usepackage{enumitem} 
\usepackage{makecell}
\usepackage{stfloats}
\usepackage{booktabs}
 \usepackage{hyperref}
\usepackage{tabularx}
\usepackage{threeparttable}
\usepackage{marvosym}
\usepackage{ifsym}
\newcommand{\partitle}[1]{\medskip \noindent \textbf{#1.}}
\newcommand{\nop}[1]{}
\AtBeginDocument{%
  \providecommand\BibTeX{{%
    \normalfont B\kern-0.5em{\scshape i\kern-0.25em b}\kern-0.8em\TeX}}}

\setcopyright{acmcopyright}
\copyrightyear{2018}
\acmYear{2018}
\acmDOI{XXXXXXX.XXXXXXX}

\acmConference[Conference acronym 'XX]{Make sure to enter the correct
  conference title from your rights confirmation emai}{June 03--05,
  2018}{Woodstock, NY}
\acmPrice{15.00}
\acmISBN{978-1-4503-XXXX-X/18/06}

\begin{document}

\title{Shapley Value on Probabilistic Classifiers}


\author{Xiang Li}
\authornote{Equal contribution.}
\affiliation{
    \institution{Zhejiang University}
    \country{}
}
\email{lixiangzx@zju.edu.cn}

\author{Haocheng Xia}
\authornotemark[1]
\affiliation{
    \institution{Zhejiang University}
    \country{}
}
\email{xiahc@zju.edu.cn}

\author{Jinfei Liu}
\authornote{Corresponding author.}
\affiliation{
    \institution{Zhejiang University}
    \country{}
}
\email{jinfeiliu@zju.edu.cn}


\begin{abstract}
Data valuation has become an increasingly significant discipline in data science due to the economic value of data. In the context of machine learning (ML), data valuation methods aim to equitably measure the contribution of each data point to the utility of an ML model. One prevalent method is Shapley value, which helps identify data points that are beneficial or detrimental to an ML model. However, traditional Shapley-based data valuation methods may not effectively distinguish between beneficial and detrimental training data points for probabilistic classifiers. In this paper, we propose Probabilistic Shapley (P-Shapley) value by constructing a probability-wise utility function that leverages the predicted class probabilities of probabilistic classifiers rather than binarized prediction results in the traditional Shapley value. We also offer several activation functions for confidence calibration to effectively quantify the marginal contribution of each data point to the probabilistic classifiers. Extensive experiments on four real-world datasets demonstrate the effectiveness of our proposed P-Shapley value in evaluating the importance of data for building a high-usability and trustworthy ML model. 
% Experiment 
\end{abstract}

\maketitle

\section{\MakeUppercase{Introduction}}\label{sec:intro}

Since data creates a steady stream of wealth, the economic value of data attracts great attention from both industry and academia. Data-driven applications, and more specifically machine learning (ML), promote data valuation to become an increasingly significant discipline in data science. In the context of ML, data valuation aims to equitably measure the contribution of each data point to the utility (i.e., performance) of an ML model. To approach the goal, many data valuation methods are developed, including Leave-One-Out score~\cite{loo}, \emph{Shapley value}~\cite{DBLP:conf/icml/GhorbaniZ19}, reinforcement learning-based value~\cite{DBLP:conf/icml/YoonAP20}, etc. Among these, Shapley value has become the most prevalent method by virtue of its unique four properties for equitable payoff allocation: \emph{balance}, \emph{symmetry}, \emph{zero element}, and \emph{additivity}~\cite{shapley1953value, DBLP:journals/tkde/Pei22}. Recent research~\cite{DBLP:conf/icml/GhorbaniZ19, betashapley, csshapley} indicates that Shapley value and its variations are effective in identifying both beneficial and detrimental data for an ML model with the demonstration of various tasks such as data selection and label noise detection.

\partitle{Motivation} Shapley-based data valuation methods depend on a utility function that assesses the value of a coalition of data points by evaluating the performance of the ML model trained on the coalition. Previous work~\cite{DBLP:conf/icml/GhorbaniZ19, betashapley, csshapley, towardseffeicientdata} has commonly defined the utility function as the prediction accuracy on a validation set $\mathcal{V}$. However, this approach may not effectively distinguish between beneficial and detrimental training data points for  probabilistic classifiers. 
Consider the following scenario: given two probabilistic classifiers $C1$ and $C2$ for binary classification where the classification threshold is $50\%$ and a validation set $\mathcal{V}$ containing two data points labeled as either $0$ or $1$. We observe that $C1$ provides predictive confidence scores (i.e., predicted class probabilities) of $\{90\%, 30\%\}$ for the correct class labels, while $C2$ provides scores of $\{60\%, 30\%\}$. 
Although $C1$ demonstrates greater predictive confidence, a significant indicator of the model's trustworthiness and reliability, in predicting the correct class label than $C2$, both classifiers have an identical prediction accuracy of $50\%$ with no discernible difference. It is therefore tempting to ask: \textit{how to effectively differentiate the utility of various probabilistic classifiers?} Moreover, although prediction accuracy provides a homogeneous smallest unit of improvement (i.e., $1 / |\mathcal{V}|$), the change in confidence resulting from the addition of new data points is generally heterogeneous. For instance, an increase in predictive confidence score from $60\%$ to $70\%$ is distinct from an increase from $90\%$ to $100\%$. In most cases, the latter is regarded as more valuable and challenging to achieve. It is therefore raising another question: \textit{how to accurately quantify the marginal contribution of various data?}

\partitle{Contribution} In this paper, we propose Probabilistic Shapley (P-Shapley) value by constructing a probability-wise utility function that effectively differentiates and quantifies the contribution of each data point to the probabilistic classifiers.

For the first question, we leverage the predicted class probabilities rather than binarized prediction results as inputs for the utility function. Using the predicted class probabilities (i.e., predictive confidence scores) of probabilistic classifiers can effectively utilize the model's confidence in its predictions. For the second question, we propose a novel solution by combining the utility function based on the predicted class probabilities with confidence calibration. Specifically, we incorporate different activation functions which tune the marginal improvements in predicted class probabilities to reflect their varying contribution to the ML models. We briefly summarize our contributions as follows.

\begin{itemize}[leftmargin=*]
   \item We identify the problem of Shapley value on probabilistic classifiers and propose Probabilistic Shapley (P-Shapley) value by constructing a probability-wise utility function.
   
    \item To effectively quantify the marginal contribution of each data point to probabilistic classifiers, we offer several activation functions for confidence calibration.

    \item Extensive experiments on four real-world datasets demonstrate the effectiveness of our proposed P-Shapley value in evaluating the importance of data for building a high-usability and trustworthy ML model.
\end{itemize}

\section{\MakeUppercase{Probabilistic Shapley Value}}\label{sec:prop}
In this section, we propose Probabilistic Shapley (P-Shapley) value. In Section \ref{subsec:pre}, we briefly overview the concept of Shapley value. Section \ref{subsec:PSV} presents the detail of P-Shapley value, including the definition of the probability-wise utility function, selection of activation functions, and P-Shapley value computation based on the truncated Monte Carlo approximation~\cite{datashapley}.

\subsection{Preliminaries}\label{subsec:pre}
% We introduce the notion of standard Shapley value in the context of FL. 
Consider a set of data points $\mathcal{N}=\{1,\ldots,n\}$. A \emph{coalition} $\mathcal{S}$ is a subset of $\mathcal{N}$ that cooperates to complete an ML task, for instance, training an ML model. A utility function $\mathcal{U}(\mathcal{S})$ $(\mathcal{S} \subseteq \mathcal{N})$ is the utility of a coalition $\mathcal{S}$ for an ML task, which is typically the prediction accuracy of the model trained on $\mathcal{S}$. The \emph{marginal contribution} of data point $i$ with respect to a coalition $\mathcal{S}$ $(i \notin \mathcal{S})$ is $\mathcal{U}(\mathcal{S}\cup \{i\})-\mathcal{U}(\mathcal{S})$. 

% Shapley~\cite{shapley1953value} laid out the fundamental requirements of fair reward allocation, including balance, symmetry, additivity, and zero element.
% \begin{itemize}
%     \item \emph{Balance.} The total payoff should be fully distributed to all data.
%     \item \emph{Symmetry.} The data points with the same marginal contribution should receive the same reward. 
%     \item \emph{Additivity.} The data value on two tasks should be the sum of the value on individual tasks.
%     \item \emph{Zero element.} The value of this data point will be assigned as 0 if it does not make any marginal contribution.
% \end{itemize}

Shapley value measures the expectation of marginal contribution by data point $i$ in all possible coalitions over $\mathcal{N}$. That is,
\begin{equation}\label{equ:SV}
  \mathcal{SV}_i=\frac{1}{n} \sum_{\mathcal{S}\subseteq \mathcal{N} \setminus \{i\}}   \binom{n-1}{|\mathcal{S}|}^{-1} \left( \mathcal{U}(\mathcal{S}\cup \{i\})-\mathcal{U}(\mathcal{S})\right) 
.
\end{equation} 

\subsection{P-Shapley Value }\label{subsec:PSV}

Most prior work ~\cite{datashapley, towardseffeicientdata, betashapley, knn} utilizes the prediction accuracy on a validation set as the utility function. We propose the first probability-wise utility function that allows us to better measure the performance and value of probabilistic classifiers. 

%{CS-Shapley\cite{csshapley} was the first to address this limitation by using class-wise accuracy as the utility function instead of overall accuracy. However, it still focuses only on predictive performance at the accuracy level.} 
%By considering predictive confidence score in predictions, our proposed approach offers a more accurate assessment of data points' true worth. Utilizing well-calibrated confidence as the utility function enables us to distinguish data points based on not just their predictive power but also the level of trust we can place in those predictions, thereby enhancing the reliability of the P-Shapley value valuations.

\partitle{Utility function with confidence calibration}
Suppose we have a set of training data points $\mathcal{N} = \{1, \ldots, n\}$. Given a binary classification task where data points are labeled as either $0$ or $1$, for any data point $i = (\bm{x}_i, y_i)$ $(y_i \in \{0, 1\})$, we need to quantify the contribution of data point $i$ to the probabilistic classifier for the binary classification task. Let $\mathcal{V}$ be the validation set. For a given data coalition $\mathcal{S}$ $(\mathcal{S} \subseteq \mathcal{N})$, the probability-wise utility function $\mathcal{U}_p(\cdot)$ is defined as follows.
\begin{equation}
\mathcal{U}_p(\mathcal{S}) = \frac{1}{|\mathcal{V}|} \sum_{j \in \mathcal{V}} \left(y_j p_j + (1-y_j)(1-p_j)\right) \mathcal{I}(y_j = \hat y_j),
\end{equation}
where $\mathcal{I}(\cdot)$ is the indicator function that returns $1$ for true condition and $0$ otherwise, $y_j$ is the ground-truth label of data point $j$ from the validation set, $\hat{y}_j$ is the label of data point $j$ predicted by the  probabilistic classifier trained on $\mathcal{S}$, and $p_j$ is the predictive confidence score that data point $j$ belongs to class $1$. In contrast, the utility function $\mathcal{U}(\cdot)$ in traditional Shapley value is typically defined as $\mathcal{U}(\mathcal{S}) = \frac{1}{|\mathcal{V}|} \sum_{j \in \mathcal{V}} \mathcal{I}(y_j = \hat y_j)$.

We essentially perform a transformation on the probabilistic classifier's utility from the prediction accuracy to the average of predictive confidence scores for correctly predicted data points in the validation set. Moreover, the increase in the utility is not linearly correlated with the predictive confidence score in most cases. As mentioned in Section \ref{sec:intro}, the increase in predictive confidence score from 90\% to 100\% is more challenging than the increase from 60\% to 70\% generally, although both represent a 10\% increase in predictive confidence score. Therefore, we incorporate activation functions into the predictive confidence score to better capture the non-linear relationship between predictive confidence score and utility.

\begin{equation}\label{equ:util}
    \mathcal{U}_p(\mathcal{S}) = \frac{1}{|\mathcal{V}|} \sum_{j \in \mathcal{V}} AF\left(y_j p_j + (1-y_j)(1-p_j)\right) \mathcal{I}(y_j = \hat y_j),
\end{equation}
where $AF(\cdot)$ is the activation function.
Based on the probability-wise utility function, we can measure the expectation of marginal contribution by data point $i$ in all possible coalitions over $\mathcal{N}$ as P-Shapley value.

%accumulate a data point's marginal contribution in all possible permutations over $\mathcal{N}$ and then compute the average result as P-Shapley value. 

\begin{equation}\label{equ:SV}
  \mathcal{PSV}_i=\frac{1}{n} \sum_{\mathcal{S}\subseteq \mathcal{N} \setminus \{i\}}   \binom{n-1}{|\mathcal{S}|}^{-1} \left( \mathcal{U}_p(\mathcal{S}\cup \{i\})-\mathcal{U}_p(\mathcal{S})\right) 
.
\end{equation} 

\begin{figure}[tb]
    \centering
    \includegraphics[width=1\linewidth]{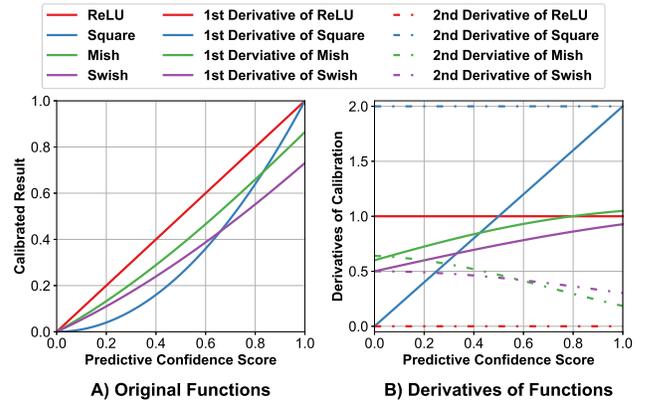}
    \vspace{-3em}
    \caption{Graphical Expressions for Activation Functions.}
    \label{fig:activate}
\end{figure}

\partitle{Activation Function Selection}
As the predictive confidence score approaches its maximum value (i.e., 100\%), achieving further improvements becomes increasingly challenging. Therefore, we aim to capture this feature and precisely calibrate the marginal contribution of marginal confidence improvement. We adopt activation functions with positive second derivatives to enhance the significance of confidence improvement for a high predictive confidence score, as shown in Figure \ref{fig:activate}. 

\begin{table}[tb]
\centering
\begin{threeparttable}
  \caption{Mathematical Expressions for Activation Functions.}
  \label{tab:act}
  \setlength{\tabcolsep}{15pt} % Default value: 6pt
  \renewcommand{\arraystretch}{1.5} % Default value: 1
  \begin{tabularx}{\linewidth}{ccl}
    \toprule
    Activation Function & Mathematical Expression\\
    \midrule
    ReLU & $y=\left\{\begin{array}{l}
        0, \text { if } x<0 \\
        x, \text { if } x \geq 0
        \end{array}\right.$\\
    Square & $y = x^2$\\
    Mish & $y = x \tanh(\ln(1+\exp(x))$ \\
    Swish* & $y = x(1+\exp(-\beta x))^{-1}$ \\
  \bottomrule
\end{tabularx}
\begin{tablenotes}
\item {\footnotesize * $\beta$ defaults to 1.}
\end{tablenotes}
\end{threeparttable}
\end{table}

The most straightforward activation function with a positive second derivative is $x^2$, which we refer to as Square. While Square possesses a monotonically increasing first derivative compared to the traditional ReLU in the range of $[0, 1]$, it causes an overly rapid value growth when the predictive confidence score is high due to its constant second derivative of 2. Therefore, we further consider popular activation functions employed in deep learning like Mish~\cite{Mish} and Swish~\cite{Swish} as shown in Table \ref{tab:act}.
To be more specific, the first derivative of Mish is 
\begin{equation}
    Mish'(x) = \omega \delta ^{-2} \exp(x), 
\end{equation}
where $\omega = 4(x + 1) + 4\exp(2x) + \exp(3x) + (4x + 6)\exp(x)$ and $\delta = 2\exp(x) + \exp(2x) + 2$. And the first derivative of Swish is
\begin{equation}
    Swish'(x) = \beta Swish(x) + (1 - \beta Swish(x))(1+\exp(-\beta x))^{-1},
\end{equation}
where $\beta$ is a trainable parameter that defaults to 1. Both the derivatives of Swish and Mish saturate beyond a threshold, differing from linear square activation derivatives. The non-linear saturating form of these activation functions' derivatives allows for a smoother value increment proportional to the predictive confidence scores, yielding more accurate calibration results for the utility function.

\partitle{P-Shapley value with truncated Monte Carlo approximation} Similar to Shapley value, calculating the exact P-Shapley value requires exponential time complexity. Therefore, we adopt an approximate truncated Monte Carlo algorithm~\cite{datashapley} to tackle the computational challenge of estimating P-Shapley value. The pseudocode is shown in Algorithm \ref{alg:tmc}. 
Specifically, we randomly sample $m$ permutations of the training set (Lines 2-3). For each permutation, we scan the data points progressively and evaluate the utility of the coalition consisting of the scanned data points (Lines 6-10). We then accumulate each data point's marginal contribution (Lines 11-12). To reduce the computational cost, we adopt a truncated threshold $\epsilon$ such that the gap between the utility of the coalition consisting of the scanned data points and the utility of the entire training set falls below $\epsilon$ (Lines 7-8). Finally, we return the average marginal contribution from all $m$ permutations as an approximation of P-Shapley value (Lines 13-14).

\begin{algorithm}[h] 
\caption{Truncated Monte Carlo for P-Shapley value.}
\label{alg:tmc}
\SetKwInOut{Input}{input}\SetKwInOut{Output}{output}
\SetCommentSty{itshape}
\Input{Training set $\mathcal{N} = \{1, \ldots, n\}$, \\ number of total permutations $m$, \\ truncated threshold $\epsilon$.}
\Output{P-Shapley value of training data points $\mathcal{PSV}_1, \dots, \mathcal{PSV}_n$.} 

    $\mathcal{PSV}_i \gets 0$ $(1 \leq i \leq n)$;\\
    \For{$t$ =1 to $m$}{
        $\pi^t \gets$ random permutation of the training set $\mathcal{N}$\;
        $\mathcal{U}_p(\emptyset) = 0$\;
        Calculate $\mathcal{U}_p(\pi^t)$ using Equation \ref{equ:util}\;
        \For{j = 1 to $n$}{
            \tcp{Denote the first j data points in $\pi^t$ as $\pi^t[:j]$}
            \If {$\mathcal{U}_p(\pi^t[:j]) - \mathcal{U}_p(\pi^t) < \epsilon$}
                {$\mathcal{U}_p(\pi^t[:j]) = \mathcal{U}_p(\pi^t[:j-1])$;}
            \Else{Calculate $\mathcal{U}_p(\pi^t[:j])$ using Equation \ref{equ:util};}
        }
        \For{$i$ = 1 to $n$}{
            $\mathcal{PSV}_i +=  \mathcal{U}_p(\pi^t[:j]) - \mathcal{U}_p(\pi^t[:j-1]);$
        }
    }
    \For{$i$ = 1 to $n$}{
        $\mathcal{PSV}_i /= m; $
    }
    \Return $\mathcal{PSV}_1, \ldots \mathcal{PSV}_n$\;
\end{algorithm}

\section{\MakeUppercase{Experiments}}
In this section, we present the empirical evaluation of the proposed algorithms on diverse classification datasets and compare their performance against existing accuracy-based data valuation methods. In Section \ref{subsec:setup},  we provide details of the experimental setup including the datasets and compared methods. In Section \ref{subsec:metric}, we propose detailed metrics for measuring predictive confidence score in the data removal experiment. In Section \ref{subsec:result}, we present and analyze several experimental results to validate the effectiveness of P-Shapley value.

\subsection{Datasets and Experimental Setup}\label{subsec:setup}
We conduct high-value data removal experiments to evaluate the effectiveness of our proposed data valuation methods. In these experiments, we iteratively remove data points from the dataset in descending order of their assessed value. Training data points with higher valuation should contribute more to the model performance, so we measure the performance of each data valuation method with the performance drop following the removal of high-value data points. % Upon each removal step, we train a new model using the updated training set and assess changes in accuracy and predictive confidence scores on a validation set. This methodology allows us to contrast the effectiveness of the data valuation methods in pinpointing the most critical data points to prune during the process.

\partitle{Compared Methods} 
We augment the proposed P-Shapley value with four different activation functions including ReLU, Square, Mish, and Swish as detailed in Table~\ref{tab:act}. We compare them with the following baseline algorithms: Leave-One-Out~\cite{loo}, truncated Monte Carlo approximated Shapley (TMC-Shapley)~\cite{datashapley}, and Beta Shapley ($\alpha=1, \beta=16$)~\cite{betashapley}. We truncate in the same iteration when estimating P-Shapley value, TMC-Shapley value, and Beta-Shapley value with the truncated Monte Carlo algorithm.

\partitle{Datasets and models}
We employ four real-world datasets from OpenML~\cite{DBLP:journals/sigkdd/VanschorenRBT13} that are commonly used to benchmark classification methods and implement a logistic regression classifier. We follow the standard methodology used in previous work~\cite{betashapley,csshapley,knn} to extract features from image datasets including Fashion-MNIST and CIFAR-10. Specifically, we utilize the pre-trained ResNet-18 model available in PyTorch~\cite{DBLP:conf/nips/PaszkeGMLBCKLGA19} to extract image representations. We then perform principal component analysis on the extracted representations and select the top 32 principal components as features.

\subsection{Evaluation Metrics}\label{subsec:metric}
\partitle{Weighted Accuracy Drop (WAD)} 
To quantify the overall accuracy drop and its rate for various data valuation methods, we adopt the weighted accuracy drop (WAD)~\cite{csshapley} as a metric. Given a training set $\mathcal{N}$ in descending order by data value and removing data points progressively starting with the highest value data point, WAD is calculated by aggregating the prediction accuracy decrease in each round, with weight inversely proportional to the number of rounds.

%WAD is calculated by cumulatively aggregating the accuracy decrease as data points are removed progressively, with the weight inversely proportional to the number of rounds.
\begin{equation}
    WAD = \sum_{j=1}^n \left( \frac{1}{j}\sum_{i=1}^j \left(ACC_{\mathcal{N}[i-1\mathbin{:}]}-ACC_{\mathcal{N}[i\mathbin{:}]}\right) \right),
\end{equation}
where $\mathcal{N}[i\mathbin{:}]$ represents the slice of $\mathcal{N}$ starting from the $i^{th}$ data point, indicating that the first $i-1$ data points have been removed. $ACC_\mathcal{N}[i\mathbin{:}]$ represents the corresponding prediction accuracy of the probabilistic classifier trained on the remaining data. For boundary cases, we define $\mathcal{N}[0\mathbin{:}]$ as the entire training set. 

\partitle{Weighted Brier Score Drop (WBD)}
In order to assess the impact on predictive confidence scores more accurately, we propose the incorporation of predicted class probabilities with weighted performance drops. Brier score is a measure of the accuracy of predicted class probabilities made by a probabilistic classifier. As in Equation \ref{equ:bs}, it is calculated as the mean squared difference between the predicted class probabilities $p_i$ and $y_i$.
\begin{equation}\label{equ:bs}
    BS = \frac{1}{n} \sum_{i=1}^n\left(y_i(p_i - y_i) + (1-y_i)p_i\right)^2.
\end{equation}
By combining the Brier score and WAD metrics, we introduce the probability-level Weighted Brier Score Drop (WBD) measure. This metric offers a probability-wise approach to evaluating model performance that considers both the effect of data point removal on model performance and its predictive confidence scores.

\begin{equation}
    WBD = -\sum_{j=1}^n \left( \frac{1}{j}\sum_{i=1}^j \left(BS_{\mathcal{N}[i-1:]}-BS_{\mathcal{N}[i:]}\right) \right).
\end{equation}

\partitle{Weighted Cross Entropy Drop (WCD)}
Similarly, we introduce cross-entropy (CE) to calculate the cumulative change in the model's predictive confidence scores.
\begin{equation}
\begin{aligned}
    CE=-\sum_{i=1}^n\left(y_i \log p_i+\left(1-y_i\right) \log \left(1-p_i\right)\right), \\
    WCD = -\sum_{j=1}^n \left( \frac{1}{j}\sum_{i=1}^j \left(CE_{\mathcal{N}[i-1:]}-CE_{\mathcal{N}[i:]}\right) \right).
\end{aligned}
\end{equation}

\subsection{Performance on Data Removal Tasks}\label{subsec:result}
\begin{table*}[t]
 \caption{Weighted Accuracy Drop, Weighted Brier Score Drop, and Weighted Cross Entropy Drop for High-value Data Removal.}
 \centering
 \begin{tabular}{lcccccccccccc}\toprule
    \setlength{\tabcolsep}{10pt} % Default value: 6pt
    & \multicolumn{3}{c}{Diabetes} & \multicolumn{3}{c}{Wind}& \multicolumn{3}{c}{Fashion-MNIST}& \multicolumn{3}{c}{CIFAR-10}
    \\\cmidrule(lr){2-4}\cmidrule(lr){5-7}\cmidrule(lr){8-10}\cmidrule(lr){11-13}
    & WAD$\uparrow$ & WBD$\uparrow$ & WCD$\uparrow$    & WAD$\uparrow$ & WBD$\uparrow$ & WCD$\uparrow$ & WAD$\uparrow$ & WBD$\uparrow$ & WCD$\uparrow$ &WAD$\uparrow$ & WBD$\uparrow$ & WCD$\uparrow$\\\midrule
    Leave-One-Out          
    & 0.154 &0.103& 0.571 
    & 0.180 &0.175& 2.283
    & 0.271 &0.181& 0.606
    & 0.109 &0.100& 0.595\\
    Beta-Shapley 
    & 0.265 &0.187& 1.521 
    & 0.274 &0.225& 5.029
    & 0.261 &0.188& 0.814
    & 0.105 &0.072& 0.291\\
    TMC-Shapley  
    & 0.414 & 0.319& 2.327 
    & 0.407 &0.340& 3.896
    & 0.380 &0.289& 1.260
    & 0.143 &0.110& 0.499\\
    P-Shapley (ReLU)     
    & 0.497 &0.380& 3.398 
    & 0.427 &0.353& 5.028
    & 0.425 &0.333& 1.530
    & 0.169 &0.134& 0.621\\
    P-Shapley (Square)    
    & \textbf{0.501} & 0.395& 3.809 
    & 0.471 &0.390& 5.148
    & 0.495 &0.387& 1.776
    & 0.260 &0.206& 1.029\\
    P-Shapley (Swish)     
    & 0.500 &\textbf{0.396}& \textbf{3.839} 
    & 0.442 &\textbf{0.397}& \textbf{5.159}
    & 0.448 &\textbf{0.399}& \textbf{1.834}
    & 0.199 &\textbf{0.213}& \textbf{1.068}\\
    P-Shapley (Mish)      
    & 0.499 &0.386& 3.451 
    & \textbf{0.479} &0.366& 5.122
    & \textbf{0.511} &0.350& 1.608
    & \textbf{0.270} &0.155& 0.741
    \\\bottomrule
 \end{tabular}
 \label{tab:datarm}
\end{table*}
\begin{figure}[ht]\label{fig:accdrop}
    \centering
    \includegraphics[width=1\linewidth]{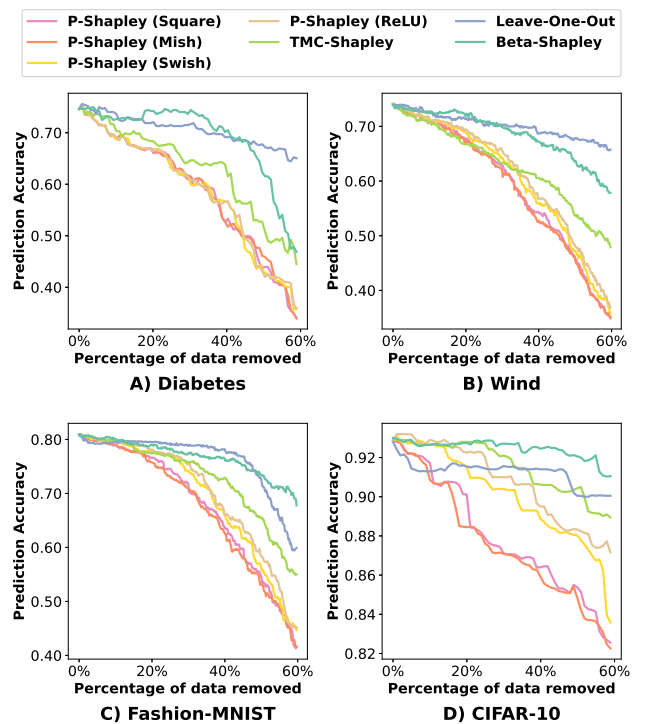}
    \caption{Results for High-value Data Removal.}
    \label{fig:datarm}
\end{figure}
Figure \ref{fig:datarm} depicts a decrease in prediction accuracy as the highest value data point is sequentially removed. The proposed P-Shapley value approach, utilizing all four activation functions, exhibits a faster decrease in accuracy as data points are removed. This indicates that P-Shapley value captures the importance of the data more precisely, allowing for more efficient data reduction. Moreover, P-Shapley value with Square, Mish, and Swish activation functions shows a faster reduction rate compared to ReLU, highlighting the efficacy of these non-linear activation functions.

Table \ref{tab:datarm} displays the reduction rates of all compared methods using the WAD, WBD, and WCD metrics, as defined in Section \ref{subsec:metric}. P-Shapley value utilizing all four activation functions consistently outperforms the baselines across all datasets. Notably, P-Shapley value with Swish activation function achieves the highest WBD and WCD scores across all datasets. One possible reason is that the Swish activation's soft clipping nature helps produce a utility function that varies smoothly with the changes in predictive confidence score, resulting in well-calibrated data valuation.

\newpage
\bibliographystyle{abbrv}
\bibliography{psv}

% %%
% %% If your work has an appendix, this is the place to put it.
% \appendix









\end{document}